\documentclass[10pt,journal,compsoc]{IEEEtran}

\usepackage{fontawesome}
\usepackage{hyperref}
\usepackage{tabularx}
\usepackage{graphicx}
\usepackage{amsmath}
\usepackage{amssymb}
\usepackage{url}
\usepackage{threeparttable}
\usepackage{framed}
\usepackage{multirow}
\usepackage{gensymb}
\usepackage{hhline}
\usepackage{rotating}
\usepackage{babel}
\usepackage{mathtools}
\usepackage{booktabs}
\usepackage{blindtext}
\usepackage[normalem]{ulem}
\usepackage{xcolor}
\usepackage{lettrine}
\usepackage{comment}
\ifCLASSOPTIONcompsoc
  \usepackage[nocompress]{cite}
\else
  \usepackage{cite}
\fi
\ifCLASSINFOpdf
\else
\fi

\let\orighref\href
\renewcommand{\href}[2]{\orighref{#1}{#2\,\faExternalLink}}

\begin{document}

\title{\emph{MedShapeNet} - A Large-Scale Dataset of 3D Medical Shapes for Computer Vision}

\author{Jianning Li, Zongwei Zhou, Jiancheng Yang, Antonio Pepe, Christina Gsaxner, Gijs Luijten, Chongyu Qu, Tiezheng Zhang, Xiaoxi Chen, Wenxuan Li, Marek Wodzinski, Paul Friedrich, Kangxian Xie, Yuan Jin, Narmada Ambigapathy, Enrico Nasca, Naida Solak, Gian Marco Melito, Viet Duc Vu, Afaque R. Memon, Christopher Schlachta, Sandrine De Ribaupierre, Rajnikant Patel, Roy Eagleson, Xiaojun Chen, Heinrich Mächler, Jan Stefan Kirschke, Ezequiel de la Rosa, Patrick Ferdinand Christ, Hongwei Bran Li, David G. Ellis, Michele R. Aizenberg, Sergios Gatidis, Thomas Küstner, Nadya Shusharina, Nicholas Heller, Vincent Andrearczyk, Adrien Depeursinge, Mathieu Hatt, Anjany Sekuboyina, Maximilian Löffler, Hans Liebl, Reuben Dorent, Tom Vercauteren, Jonathan Shapey, Aaron Kujawa, Stefan Cornelissen, Patrick Langenhuizen, Achraf Ben-Hamadou, Ahmed Rekik, Sergi Pujades, Edmond Boyer, Federico Bolelli, Costantino Grana, Luca Lumetti, Hamidreza Salehi, Jun Ma, Yao Zhang, Ramtin Gharleghi, Susann Beier, Arcot Sowmya, Eduardo A. Garza-Villarreal, Thania Balducci, Diego Angeles-Valdez, Roberto Souza, Leticia Rittner, Richard Frayne, Yuanfeng Ji, Vincenzo Ferrari, Soumick Chatterjee, Florian Dubost, Stefanie Schreiber, Hendrik Mattern, Oliver Speck, Daniel Haehn, Christoph John, Andreas Nürnberger, João Pedrosa, Carlos Ferreira, Guilherme Aresta, António Cunha, Aurélio Campilho, Yannick Suter, Jose Garcia, Alain Lalande, Vicky Vandenbossche, Aline Van Oevelen, Kate Duquesne, Hamza Mekhzoum, Jef Vandemeulebroucke, Emmanuel Audenaert, Claudia Krebs, Timo van Leeuwen, Evie Vereecke, Hauke Heidemeyer, Rainer Röhrig, Frank Hölzle, Vahid Badeli, Kathrin Krieger, Matthias Gunzer, Jianxu Chen, Timo van Meegdenburg, Amin Dada, Miriam Balzer, Jana Fragemann, Frederic Jonske, Moritz Rempe, Stanislav Malorodov, Fin H. Bahnsen, Constantin Seibold, Alexander Jaus, Zdravko Marinov, Paul F. Jaeger, Rainer Stiefelhagen, Ana Sofia Santos, Mariana Lindo, André Ferreira, Victor Alves, Michael Kamp, Amr Abourayya, Felix Nensa, Fabian Hörst, Alexander Brehmer, Lukas Heine, Yannik Hanusrichter, Martin Weßling, Marcel Dudda, Lars E. Podleska, Matthias A. Fink, Julius Keyl, Konstantinos Tserpes, Moon-Sung Kim, Shireen Elhabian, Hans Lamecker, Dženan Zukić, Beatriz Paniagua, Christian Wachinger, Martin Urschler, Luc Duong, Jakob Wasserthal, Peter F. Hoyer, Oliver Basu, Thomas Maal, Max J. H. Witjes, Gregor Schiele, Ti-chiun Chang, Seyed-Ahmad Ahmadi, Ping Luo, Bjoern Menze, Mauricio Reyes, Thomas M. Deserno, Christos Davatzikos, Behrus Puladi, Pascal Fua, Alan L. Yuille, \\ Jens Kleesiek, Jan Egger

\IEEEcompsocitemizethanks{
\IEEEcompsocthanksitem J. Li, G. Luijten, N. Ambigapathy, E. Nasca, A. Dada, M. Balzer, J. Fragemann, F. Jonske, M. Rempe, A. Abourayya, S. Malorodov, F. H. Bahnsen, C. Seibold, A. S. Santos, M. Lindo, A. Ferreira, F. Nensa, F. Hörst, A. Brehmer, L. Heine, J. Keyl, M.-S. Kim, M. Kamp, J. Kleesiek and J. Egger are with the Institute for Artificial Intelligence in Medicine (IKIM), University Hospital Essen (AöR), Girardetstraße 2, 45131 Essen, Germany. E-mails: Jianning.Li@uk-essen.de; Jan.Egger@uk-essen.de \\ \\
\IEEEcompsocthanksitem J. Li, A. Pepe, C. Gsaxner, Y. Jin, G. Luijten, N. Solak and J. Egger are with the Institute of Computer Graphics and Vision (ICG), Graz University of Technology, Inffeldgasse 16c, 8010 Graz, Austria.
\IEEEcompsocthanksitem J. Li, A. Pepe, C. Gsaxner, Y. Jin, G. Luijten, N. Solak and J. Egger are with Computer Algorithms for Medicine Laboratory (Cafe), Graz, Austria.

\IEEEcompsocthanksitem Z. Zhou, C. Qu, T. Zhang, W. Li, and A. L. Yuille are with the Department of Computer Science, Johns Hopkins University, Malone Hall, 3400 N Charles St, Baltimore, MD 21218, USA.

\IEEEcompsocthanksitem J. Yang and P. Fua are with the Computer Vision Laboratory, Swiss Federal Institute of Technology Lausanne (EPFL), Rte Cantonale, Lausanne 1015, Switzerland.

\IEEEcompsocthanksitem M. Wodzinski is with the Department of Measurement and Electronics, AGH University of Science and Technology, Krakow, Poland and the Information Systems Institute, University of Applied Sciences Western Switzerland (HES-SO Valais), Sierre, Switzerland.

\IEEEcompsocthanksitem P. Friedrich is with the Center for medical Image Analysis \& Navigation (CIAN),  Department of Biomedical Engineering, University of Basel, Hegenheimermattweg 167C CH-4123 Allschwil, Switzerland.

\IEEEcompsocthanksitem K. Xie is with the Boston College, 140 Commonwealth Ave, Chestnut Hill, MA 02467, USA.

\IEEEcompsocthanksitem C. Schlachta, S. De Ribaupierre, R. Patel and R. Eagleson are with Canadian Surgical Technologies \& Advanced Robotics (CSTAR), University Hospital, B7-200, 339 Windermere Road, London, N6A 5A5, Canada.

\IEEEcompsocthanksitem X. Chen is with the Department of Radiology, Renji Hospital, School of Medicine, Shanghai Jiao Tong University, Shanghai, 200240, China.

\IEEEcompsocthanksitem Y. Jin is with the Research Center for Connected Healthcare Big Data, ZhejiangLab, Hangzhou, Zhejiang, 311121 China.
\IEEEcompsocthanksitem T. van Meegdenburg is with the Faculty of Statistics, Technical University Dortmund, August-Schmidt-Straße 1, 44227 Dortmund, Germany and the Institute for Artificial Intelligence in Medicine (IKIM), University Hospital Essen (AöR), Girardetstraße 2, 45131 Essen, Germany.
\IEEEcompsocthanksitem G. M. Melito is with the Institute of Mechanics, Graz University of Technology, Kopernikusgasse 24/IV, 8010 Graz, Austria.
\IEEEcompsocthanksitem X. Chen is with the Institute of Biomedical Manufacturing and Life Quality Engineering, State Key Laboratory of Mechanical System and Vibration, School of Mechanical Engineering, Shanghai Jiao Tong University, 800 Dongchuan Road, Shanghai, 200240, People’s Republic of China.
\IEEEcompsocthanksitem H. Mächler is with the Department of Cardiac Surgery, Medical University Graz, Auenbruggerplatz 29, 8036 Graz, Austria.
\IEEEcompsocthanksitem V. D. Vu is with the Department of Diagnostic and Interventional Radiology, University Hospital Giessen, Justus-Liebig-University Giessen, Klinikstraße 33, 35392 Giessen, Germany.
\IEEEcompsocthanksitem A. R. Memon is with the Department of Mechanical Engineering, Mehran University of Engineering and Technology, Jamshoro 76062, Sindh, Pakistan.
\IEEEcompsocthanksitem A. R. Memon and X. Chen are with the Institute of Medical Robotics, Shanghai Jiao Tong University, Shanghai, People’s Republic of China.
\IEEEcompsocthanksitem J. S. Kirschke is with the Geschäftsführender Oberarzt Abteilung für Interventionelle und Diagnostische Neuroradiologie, Universitätsklinikum der Technischen Universität München, Ismaningerstr. 22, 81675 München, Germany.
\IEEEcompsocthanksitem Ezequiel de la Rosa is with icometrix, Kolonel Begaultlaan 1b, 3012 Leuven, Belgium, and the Department of Informatics, Technical University of Munich, Boltzmannstraße 3, 85748 Garching bei München, Germany.
\IEEEcompsocthanksitem S. Gatidis and T. Küstner are with the University Hospital of Tuebingen Diagnostic and Interventional Radiology Medical Image and Data Analysis (MIDAS.lab), Otfried-Müller-Str. 3, 72016 Tuebingen, Germany.
\IEEEcompsocthanksitem D. G. Ellis and M. R. Aizenberg are with the Department of Neurosurgery, University of Nebraska Medical Center, Omaha, NE, 68198 USA.
\IEEEcompsocthanksitem N. Shusharina is with the Division of Radiation Biophysics, Department of Radiation Oncology, Massachusetts General Hospital and Harvard Medical School, 55 Fruit St, Boston, Massachusetts 02114 USA.
\IEEEcompsocthanksitem N. Heller is with the University of Minnesota, Minneapolis, MN 55455 USA.
\IEEEcompsocthanksitem V. Andrearczyk and A. Depeursinge are with the Institute of Informatics, HES-SO Valais-Wallis University of Applied Sciences and Arts Western Switzerland, rue du Technopole 3, 3960 Sierre, Switzerland; A. Depeursinge is also with the Department of Nuclear Medicine and Molecular Imaging, Lausanne University Hospital (CHUV), Rue du Bugnon 46, 1005 Lausanne, Switzerland.
\IEEEcompsocthanksitem M. Hatt is with LaTIM, INSERM, UMR 1101, Univ Brest, Brest, France.
\IEEEcompsocthanksitem R. Dorent, T. Vercauteren, J. Shapey and A. Kujawa are with the King's College London, Strand, London WC2R 2LS, UK; R. Dorent is also with the Department of Neurosurgery, Brigham and Women’s Hospital, Harvard Medical School, 75 Francis St, Boston, MA 02115 USA.
\IEEEcompsocthanksitem S. Cornelissen and P. Langenhuizen are with Elisabeth-TweeSteden Hospital, Hilvarenbeekse Weg 60, 5022 GC Tilburg, Netherlands and the Video Coding \& Architectures Research Group, Department of Electrical Engineering, Eindhoven University of Technology, Groene Loper 3, 5612 AE, Eindhoven, Netherlands.
\IEEEcompsocthanksitem A. B. Hamadou and A. Rekik are with the Centre de Recherche en Num\'{e}rique de Sfax, Laboratory of Signals, Systems, Artificial Intelligence and Networks, Technop\^{o}le de Sfax, 3021 Sfax, Tunisia, and Udini, 37 BD Aristide Briand, 13100 Aix-En-Provence, France.
\IEEEcompsocthanksitem S. Pujades and E. Boyer are with Inria, Université Grenoble Alpes, CNRS, Grenoble INP, LJK, 38000 Grenoble, France.

\IEEEcompsocthanksitem A. Sekuboyina is with the Department of Informatics, Technical University of Munich, Germany, Boltzmannstraße 3, 85748 Garching bei München, Germany.
\IEEEcompsocthanksitem Maximilian Löffler is with the Universitätsklinikum Freiburg, Hugstetter Strasse 55, 79106 Freiburg, Germany.
\IEEEcompsocthanksitem H. Liebl is with the Department of Neuroradiology, Klinikum Rechts der Isar, Ismaninger Str. 22, 81675 Munich, Germany.
\IEEEcompsocthanksitem P. F. Christ, H. B. Li and B. Menze are with the Department of Quantitative Biomedicine, University of Zurich, Winterthurerstrasse 190, 8057 Zurich, Switzerland.
\IEEEcompsocthanksitem F. Bolelli, C. Grana and L. Lumetti are with the University of Modena and Reggio Emilia, Department of Engineering "Enzo Ferrari", Via Vivarelli 10, 41125, Modena, Italy.
\IEEEcompsocthanksitem J. Ma is with the Department of Laboratory Medicine and Pathobiology, University of Toronto, Toronto, ON M5S 1A8 Canada; Peter Munk Cardiac Centre, University Health Network, 585 University Ave, Toronto, ON M5G 2N2, Canada; Vector Institute, 661 University Ave Suite 710, Toronto, ON M5G 1M1, Canada.
\IEEEcompsocthanksitem Y. Zhang is with the Shanghai AI Laboratory, Yunjin Road, Shanghai, 200032, People’s Republic of China.
\IEEEcompsocthanksitem R. Gharleghi and S. Beier are with the School of Mechanical and Manufacturing Engineering, UNSW, Sydney, 2052, NSW, Australia.
\IEEEcompsocthanksitem A. Sowmya is with the School of Computer Science and Engineering, UNSW, Sydney, 2052, NSW, Australia.
\IEEEcompsocthanksitem R. Souza is with the Advanced Imaging and Artificial Intelligence Lab, Electrical and Software Engineering Department, and the Hotchkiss Brain Institute, University of Calgary, Calgary, Canada.
\IEEEcompsocthanksitem L. Rittner is with the Medical Image Computing Lab, School of Electrical and Computer Engineering (FEEC), University of Campinas, Campinas, Brazil.
\IEEEcompsocthanksitem R. Frayne is with the Radiology and Clinical Neurosciences Departments, the Hotchkiss Brain Institute, University of Calgary, Calgary, Canada, and the Seaman Family MR Research Centre, Foothills Medical Center, Calgary, Canada.
\IEEEcompsocthanksitem T.-c. Chang is with Merck, Rahway, NJ 07065, USA.
\IEEEcompsocthanksitem S.-A. Ahmadi is with the NVIDIA GmbH, Bavaria Towers - Blue Tower, Einsteinstrasse 172, 81677 Munich, Germany.
\IEEEcompsocthanksitem Y. Ji and P. Luo are with the University of Hongkong, Pok Fu Lam, Hong Kong, People’s Republic of China.
\IEEEcompsocthanksitem H. Salehi is with the Department of Artificial Intelligence in Medical Sciences, Faculty of Advanced Technologies in Medicine, Iran University Of Medical Sciences, Tehran, Iran.
\IEEEcompsocthanksitem J. Pedrosa, C. Ferreira, A. Cunha and A. Campilho are with the Institute for Systems and Computer Engineering, Technology and Science (INESC TEC), Porto, Portugal; J. Pedrosa, C. Ferreira and A. Campilho are also with the Faculty of Engineering of the University of Porto (FEUP), Porto, Portugal. A. Cunha is also with the Universidade of Trás os Montes and Alto Douro (UTAD), Vila Real, Portugal.
\IEEEcompsocthanksitem G. Aresta is with the Christian Doppler Lab for Artificial Intelligence in Retina, Department of Ophthalmology and Optometry, Medical University of Vienna, Austria.
\IEEEcompsocthanksitem Y. Suter and M. Reyes are with the ARTORG Center for Biomedical Engineering Research, University of Bern, Bern, Switzerland. M. Reyes is also with the Department of Radiation Oncology, University Hospital Bern, University of Bern, Switzerland.

\IEEEcompsocthanksitem T. M. Deserno is with the Peter L. Reichertz Institute for Medical Informatics of TU Braunschweig and Hannover Medical School, M\"uhlenpfordtstr. 23, 38106 Braunschweig, Germany.

\IEEEcompsocthanksitem J. Garcia is with the Center for Biomedical Image Computing and Analytics (CBICA), Perelman School of Medicine, University of Pennsylvania.
\IEEEcompsocthanksitem A. Lalande is with the ICMUB laboratory, CNRS UMR 6302 Faculty of Medicine, University of Burgundy, 7 Bld Jeanne d’Arc, BP 87900, 21079 Dijon, cedex, France and Medical Imaging Department - University Hospital of Dijon, 1 Bld Jeanne d’Arc, BP 77908, 21079 Dijon Cedex, France.
\IEEEcompsocthanksitem V. Vandenbossche, A. Van Oevelen and K. Duquesne are with the Department of Human Structure and Repair, Ghent University, Corneel Heymanslaan 10, 9000 Ghent, Belgium.
\IEEEcompsocthanksitem H. Mekhzoum and J. Vandemeulebroucke are with the Department of Electronics and Informatics (ETRO), Vrije Universiteit Brussel, Brussels, Belgium.
\IEEEcompsocthanksitem E. Audenaert is with the Department of Human Structure and Repair, Ghent University, Corneel Heymanslaan 10, 9000 Ghent, Belgium.
\IEEEcompsocthanksitem C. Krebs is with the Department of Cellular and Physiological Sciences, Life Sciences Centre, 1544 - 2350 Health Sciences Mall, University of British Columbia, Vancouver, British Columbia, V6T 1Z3 Canada.
\IEEEcompsocthanksitem E. Vereecke and T. V. Leeuwen are with the Department of Development \& Regeneration, KU Leuven Campus Kulak, Etienne Sabbelaan 53, 8500 Kortrijk, Belgium.
\IEEEcompsocthanksitem M.-S. Kim and F. Nensa are with the Institute of Diagnostic and Interventional Radiology and Neuroradiology, University Hospital Essen (AöR), Hufelandstraße 55, 45147 Essen, Germany.
\IEEEcompsocthanksitem J. Kleesiek is with the German Cancer Consortium (DKTK), Partner Site Essen, Hufelandstraße 55, 45147 Essen, Germany, and the Department of Physics, TU Dortmund University, August-Schmidt-Str. 4, 44227 Dortmund, Germany.
\IEEEcompsocthanksitem M. Kamp, F. Hörst, M.-S. Kim, J. Kleesiek and J. Egger are with the Cancer Research Center Cologne Essen (CCCE), University Medicine Essen (AöR), Hufelandstraße 55, 45147 Essen, Germany.
\IEEEcompsocthanksitem M. Kamp and A. Abourayya are with the Institute for Neuroinformatics, Ruhr University Bochum, Germany. M. Kamp is also with the Department of Data Science \& AI, Monash University, Australia.
\IEEEcompsocthanksitem C. Gsaxner, F. Hölzle and B. Puladi are with the Department of Oral and Maxillofacial Surgery, University Hospital RWTH Aachen, Pauwelsstraße 30, 52074 Aachen, Germany.
\IEEEcompsocthanksitem V. Badeli is with the Institute of Fundamentals and Theory in Electrical Engineering, Graz University of Technology, 8010 Graz, Austria.
\IEEEcompsocthanksitem K. Krieger, M. Gunzer and J. Chen are with the Leibniz-Institut für Analytische Wissenschaften-ISAS-e.V., 44139 Dortmund, Germany.
\IEEEcompsocthanksitem M. Gunzer is with the Institute for Experimental Immunology and Imaging, University Hospital, University Duisburg-Essen, Hufelandstrasse 55, 45147 Essen, Germany.
\IEEEcompsocthanksitem H. Heidemeyer, R. Röhrig and B. Puladi are with the Institute of Medical Informatics, University Hospital RWTH Aachen, Pauwelsstraße 30, 52074 Aachen, Germany.

\IEEEcompsocthanksitem A. Jaus, Z. Marinov and R. Stiefelhagen are with the Computer Vision for Human-Computer Interaction Lab, Department of Informatics, Karlsruhe Institute of Technology, Adenauerring 10, 76131 Karlsruhe, Germany.

\IEEEcompsocthanksitem P. F. Jaeger is with the German Cancer Research Center (DKFZ) Heidelberg, Interactive Machine Learning Group, Neuenheimer Feld 223, 69120 Heidelberg, Germany, and Helmholtz Imaging, DKFZ Heidelberg, Neuenheimer Feld 223, 69120 Heidelberg, Germany.

\IEEEcompsocthanksitem A. S. Santos, M. Lindo, A. Ferreira and V. Alves are with the Center Algoritmi / LASI, University of Minho, Braga, 4710-057, Portugal.

\IEEEcompsocthanksitem Y. Hanusrichter and M. Weßling are with the Department of Tumour Orthopaedics and Revision Arthroplasty, Orthopaedic Hospital Volmarstein, Lothar-Gau-Str. 11, 58300 Wetter, Germany, and the Center for Musculoskeletal Surgery, University Hospital of Essen, Hufelandstr. 55, 45147 Essen, Germany.

\IEEEcompsocthanksitem M. Dudda is with the Department of Trauma, Hand and Reconstructive Surgery, University Hospital Essen, Hufelandstr. 55, 45147 Essen, Germany.

\IEEEcompsocthanksitem L. E. Podleska is with the Department of Tumor Orthopedics and Sarcoma Surgery, University Hospital Essen (AöR), Hufelandstraße 55, 45147 Essen, Germany.

\IEEEcompsocthanksitem M. A. Fink is with the Clinic for Diagnostic and Interventional Radiology, University Hospital Heidelberg, Im Neuenheimer Feld 420, 69120 Heidelberg, Germany.
\IEEEcompsocthanksitem K. Tserpes is with the Department of Informatics and Telematics, Harokopio University of Athens, 9 Omirou, 177 78, Tavros, Greece.
\IEEEcompsocthanksitem E. A. Garza-Villarreal, T. Balducci and D. Angeles-Valdez are with the Institute of Neurobiology, Universidad Nacional Autónoma de México campus Juriquilla, Boulevard Juriquilla 3001, Juriquilla, Querétaro, 76230, México. D. Angeles-Valdez is also with the Department of Biomedical Sciences of Cells and Systems, Cognitive Neuroscience Center, University Medical Center Groningen, University of Groningen, Hanzeplein 1, 9713 GZ Groningen, Netherlands.

\IEEEcompsocthanksitem F. Dubost is with Google LLC, California, USA.
\IEEEcompsocthanksitem S. Schreiber, H. Mattern and O. Speck are with the German Centre for Neurodegenerative Disease, Magdeburg, Germany; and Centre for Behavioural Brain Sciences, Magdeburg, Germany.

\IEEEcompsocthanksitem S. Schreiber is also with the Department of Neurology, Medical Faculty, University Hospital Magdeburg, Germany.
\IEEEcompsocthanksitem H. Mattern is also with the Department of Biomedical Magnetic Resonance, Institute for Physics, Faculty of Natural Sciences, Otto von Guericke University Magdeburg, Germany.
\IEEEcompsocthanksitem O. Speck is also with the Department of Biomedical Magnetic Resonance, Institute for Physics, Faculty of Natural Sciences, Otto von Guericke University Magdeburg, Germany.

\IEEEcompsocthanksitem D. Haehn is with the University of Massachusetts Boston 100 Morrissey Blvd, Boston, MA 02125, USA.

\IEEEcompsocthanksitem C. John is with Christoph John - Ecubed Solutions, Berliner Ring 97, 64625 Bensheim, Germany.

\IEEEcompsocthanksitem V. Ferrari is with the Dipartimento di Ingegneria dell’Informazione, University of Pisa, Pisa, Italy and EndoCAS Center, Department of Translational Research and of New Surgical and Medical Technologies, University of Pisa, Pisa, Italy.

\IEEEcompsocthanksitem S. Chatterjee and A. Nürnberger are with Data and Knowledge Engineering Group, Faculty of Computer Science, Otto von Guericke University Magdeburg, Universitätspl. 2, 39106 Magdeburg, Germany. S. Chatterjee is also with Genomics Research Centre, Human Technopole, Milan, Italy. A. Nürnberger is also with Centre for Behavioural Brain Sciences, Magdeburg, Universitätspl 2,
39106 Magdeburg, Germany.
\IEEEcompsocthanksitem S. Elhabian is with the Scientific Computing and Imaging Institute, University of Utah, Utah, 72 South Central Campus Drive, Salt Lake City, 84112 USA.
\IEEEcompsocthanksitem H. Lamecker is with 1000shapes GmbH, Hamerlingweg 5, 14167 Berlin, Germany.
\IEEEcompsocthanksitem B. Paniagua and Dž. Zukić are with Medical Computing, Kitware Inc., Carrboro, North Carolina, 101 East Weaver St, Suite G4, 27510 USA.
\IEEEcompsocthanksitem C. Wachinger is with the Lab for Artificial Intelligence in Medical Imaging, Department of Radiology, Technical University Munich, Ismaningerstr. 22, 81675 Munich, Germany.
\IEEEcompsocthanksitem M. Urschler is with the Institute for Medical Informatics, Statistics and Documentation, Medical University Graz, Auenbruggerplatz 2, 8036 Graz, Austria.
\IEEEcompsocthanksitem J. Wasserthal is with the Clinic of Radiology \& Nuclear Medicine, University Hospital Basel, Petersgraben 4, 4031 Basel, Switzerland.
\IEEEcompsocthanksitem L. Duong is with the Département de génie logiciel et des TI, École de technologie supérieure, 1100 Notre-Dame Ouest, Montréal (Québec) H3C 1K3, Canada.
\IEEEcompsocthanksitem C. Davatzikos is with the Center for Biomedical Image Computing and Analytics, Penn Neurodegeneration Genomics Center, and Center for AI And Data Science For Integrated Diagnostics, University of Pennsylvania, Philadelphia, PA 19104 USA.
\IEEEcompsocthanksitem P. F. Hoyer is with the Department of Pediatrics II, University Hospital Essen, Hufelandstraße 55, 45147 Essen, Germany.
\IEEEcompsocthanksitem O. Basu is with the Department of Pediatrics III, University Hospital Essen, University Medicine Essen, Hufelandstraße 55, 45147 Essen, Germany.
\IEEEcompsocthanksitem T. Maal is with the 3D Imaging Lab, Radboud University Nijmegen Medical Centre, Geert Grooteplein 10, 6525 GA, Nijmegen, The Netherlands.
\IEEEcompsocthanksitem M. J. H. Witjes is with the Department of Oral and Maxillofacial Surgery, University of Groningen, University Medical Center Groningen, Hanzeplein 1 BB70, 9713 GX Groningen, The Netherlands.

\IEEEcompsocthanksitem G. Schiele is with the Intelligent Embedded Systems Lab, University of Duisburg-Essen, Bismarckstraße 90, 47057 Duisburg, Germany.

\IEEEcompsocthanksitem O. Basu and J. Egger are with the Center for Virtual and Extended Reality in Medicine (ZvRM), University Hospital Essen, University Medicine Essen, Hufelandstraße 55, 45147 Essen, Germany.
}
}

\IEEEtitleabstractindextext{%
\begin{abstract}
 Prior to the deep learning era, \textit{shape} was commonly used to describe the objects. Nowadays, state-of-the-art (SOTA) algorithms in medical imaging are predominantly diverging from computer vision, where voxel grids, meshes, point clouds, and implicit surface models are used. This is seen from numerous shape-related publications in premier vision conferences as well as the growing popularity of \textit{ShapeNet} (about 51,300 models) and \textit{Princeton ModelNet} (127,915 models). For the medical domain, we present a large collection of anatomical shapes (e.g., bones, organs, vessels) and 3D models of surgical instrument, called \textit{MedShapeNet}, created to facilitate the translation of data-driven vision algorithms to medical applications and to adapt SOTA vision algorithms to medical problems. As a unique feature, we directly model the majority of shapes on the imaging data of real patients. As of today, \textit{MedShapeNet} includes 23 datasets with more than 100,000 shapes that are paired with annotations (ground truth). Our data is freely accessible via a web interface and a Python application programming interface (API) and can be used for discriminative, reconstructive, and variational benchmarks as well as various applications in virtual, augmented, or mixed reality, and 3D printing. Exemplary, we present use cases in the fields of classification of brain tumors, skull reconstructions, multi-class anatomy completion, education, and 3D printing. In future, we will extend the data and improve the interfaces. 
 The project pages are: \url{https://medshapenet.ikim.nrw/} and \url{https://github.com/Jianningli/medshapenet-feedback}.
\end{abstract}

}

\maketitle

\IEEEdisplaynontitleabstractindextext
\IEEEpeerreviewmaketitle


\section{Introduction}
\lettrine[findent=2pt]{\textbf{T}}{}he success of deep learning in many fields of applications, including vision \cite{esteva2021deep}, language \cite{young2018recent} and speech \cite{latif2020deep}, is mainly due to the availability of large, high-quality datasets \cite{sun2017revisiting, EGGER_MedicalMeta,egger2021deep}, such as \textit{ImageNet} \cite{deng2009imagenet}, \textit{CIFAR} \cite{krizhevsky2009learning}, \textit{Penn Treebank} \cite{taylor2003penn}, \textit{WikiText} \cite{merity2016pointer} and \textit{LibriSpeech} \cite{panayotov2015librispeech}. In 3D computer vision, \textit{Princeton ModelNet} \cite{wu20153d}, \textit{ShapeNet} \cite{chang2015shapenet}, etc., are the de facto benchmarks for numerous fundamental vision problems, including 3D shape classification and retrieval \cite{lin2021single}, shape completion \cite{yan2022shapeformer}, shape reconstruction and segmentation \cite{yi2017large}. Shape describes the geometries of 3D objects and is one of the most basic concepts in computer vision. Common 3D shape representations include point clouds, voxel occupancy grids, meshes, and implicit surface models (signed distance functions), which follow different data structures, cater for different algorithms, and are convertible to each other \cite{sarasua2022hippocampal}. These shape representations diverge from gray-scale medical imaging data routinely used in clinical diagnosis and treatment procedures, such as computed tomography (CT), magnetic resonance imaging (MRI), positron emission tomography (PET), ultra sound (US), and X-ray.

The concept of shape in medical imaging is not novel. For example, statistical shape modeling (SSM) has been a longstanding method for medical image segmentation \cite{heimann2009statistical} and 3D anatomy modeling \cite{petrelli2022geometric}. The use of shape priors and constraints can also benefit medical image segmentation and reconstruction tasks \cite{yang2022implicitatlas}. Furthermore, the prominent \textit{Medical Image Computing and Computer Assisted Intervention (MICCAI)} society has established a special interest group in \textit{Shape in Medical Imaging} (\textit{ShapeMI}). This group is dedicated to exploring the applications of both traditional and contemporary (e.g., learning-based) shape analysis methods in medical imaging. Table \ref{table:shapemi} presents a partial list of professional organizations and events that are committed to this objective.

Nevertheless, state-of-the-art (SOTA) algorithms connot be directly applied to medical problems, since the vision methods were developed on general 3D shapes from \textit{ShapeNet} and not on volumetric, gray-scale medical data. Therefore, the community needs a large, high-quality shape database for medical imaging that represents a variety of 3D medical shapes, i.e., voxel occupancy grid (VOR), mesh and point representations of human anatomies \cite{rezanejad2022medial}. The inclusion of diverse anatomical shapes can aid in the development and evaluation of data-driven, shape-based methods for both vision and medical problems. 

Computer vision methods, such as facial modeling \cite{kania2023blendfields} and internal anatomy inference \cite{keller2022osso} involve anatomical shapes, and medical problems can be solved using shape-based methods. Cranial implant design \cite{li2020baseline,morais2019automated,li2021autoimplant,li2021automatic,li2023towards} is a typical example of a clinical problem that is commonly solved using well-established shape completion methods \cite{dai2017shape}. Such a shape completion concept can also be straightforwardly extended to other anatomical structures or even the whole body \cite{li2023completor}. Therefore, there is a need for both normal and pathological anatomies to solve shape-based problems that are conventionally addressed using gray-scale medical images, e.g., extacting biomarkers \cite{zhang2020automatic}.

\begin{table*}[ht]
\centering
\scriptsize
\caption{A Non-inclusive List of Organizations \& Events Featuring Shape and Computer Vision Methods for Medical Applications}
\begin{tabular}[t]{lll} 
\toprule
Sources (link) & Description & Category\\
\hline
\textit{Zuse Institute Berlin (ZIB)} \href{https://www.zib.de/projects/shape-informed-medical-image-segmentation}{}, \href{https://www.zib.de/projects/geometric-analysis-human-spine-image-based-diagnosis-biomechanical-analysis-and-neurosurgery}{} & shape-informed medical image segmentation and shape priors in medical imaging & research group\\
\textit{ShapeMI} \href{https://shapemi.github.io/}{} & shape processing/analysis/learning in medical imaging & MICCAI workshop\\
\textit{SIG} \href{http://www.miccai.org/special-interest-groups/shape-modeling-and-analysis/}{} & shape modeling and analysis in medical imaging & MICCAI special interest group (SIG) \\
\textit{AutoImplant I, II} \href{https://autoimplant.grand-challenge.org/}{}, \href{https://autoimplant2021.grand-challenge.org/}{} & skull shape reconstruction and completion & MICCAI challenge\\
\textit{WiSh} \href{https://www.birs.ca/events/2021/5-day-workshops/21w5128}{} & women in Shape Analysis, shape modeling & professional organization\\
\textit{STACOM} \href{https://link.springer.com/book/10.1007/978-3-030-93722-5}{} & statistical atlases and computational models of the heart & MICCAI workshop \\
\textit{SAMIA} \href{https://link.springer.com/book/10.1007/978-3-319-03813-1}{} & shape analysis in medical image analysis & book \\
\textit{CIBC} \href{https://www.sci.utah.edu/cibc-trd/160-iga.html}{} & image and geometric analysis &research group \\
\textit{GeoMedIA} \href{https://geomedia-workshop.github.io/}{} & geometric deep learning in medical image analysis & MICCAI-endorsed workshop\\
\textit{IEEE TMI} \href{https://www.embs.org/wp-content/uploads/2021/02/TMI-CFP-GDL_final.pdf}{} & geometric deep learning in medical imaging & journal special issue \\
\textit{PMLR} \href{https://proceedings.mlr.press/v194/}{} & geometric deep learning in medical image analysis & proceedings\\
\textit{Elsevier} \href{}{} & Riemannian geometric statistics in medical image analysis & book\\
\textit{Springer} \href{https://link.springer.com/book/10.1007/978-3-642-55987-7}{} & geometric methods in bio-medical image processing & proceedings\\
\textit{MCV} \href{https://mcv-workshop.github.io/}{}  & workshop on medical computer vision & CVPR workshop\\
\textit{MCV 2010 - 2016} \href{https://link.springer.com/conference/mcv}{}  & workshop on medical computer vision & MICCAI workshop \\ 
\textit{MeshMed} \href{https://link.springer.com/book/10.1007/978-3-642-33463-4}{}  & workshop on mesh processing in medical image analysis & MICCAI workshop \\ 
\bottomrule
\end{tabular}
\label{table:shapemi}
\end{table*}

In this paper, we present \textit{MedShapeNet},  
(1) a unique dataset for medical imaging shapes that serve complementary to existing shape benchmarks in computer vision, (2) a gap-bridger between the medical imaging and computer vision communities, and (3) a publicly available, continuous extending resource for benchmarking,  education, extended reality (XR) applications \cite{gsaxner2021inside}, and the investigation of anatomical shape variations.

\begin{figure}
\centering
\includegraphics[width=\linewidth]{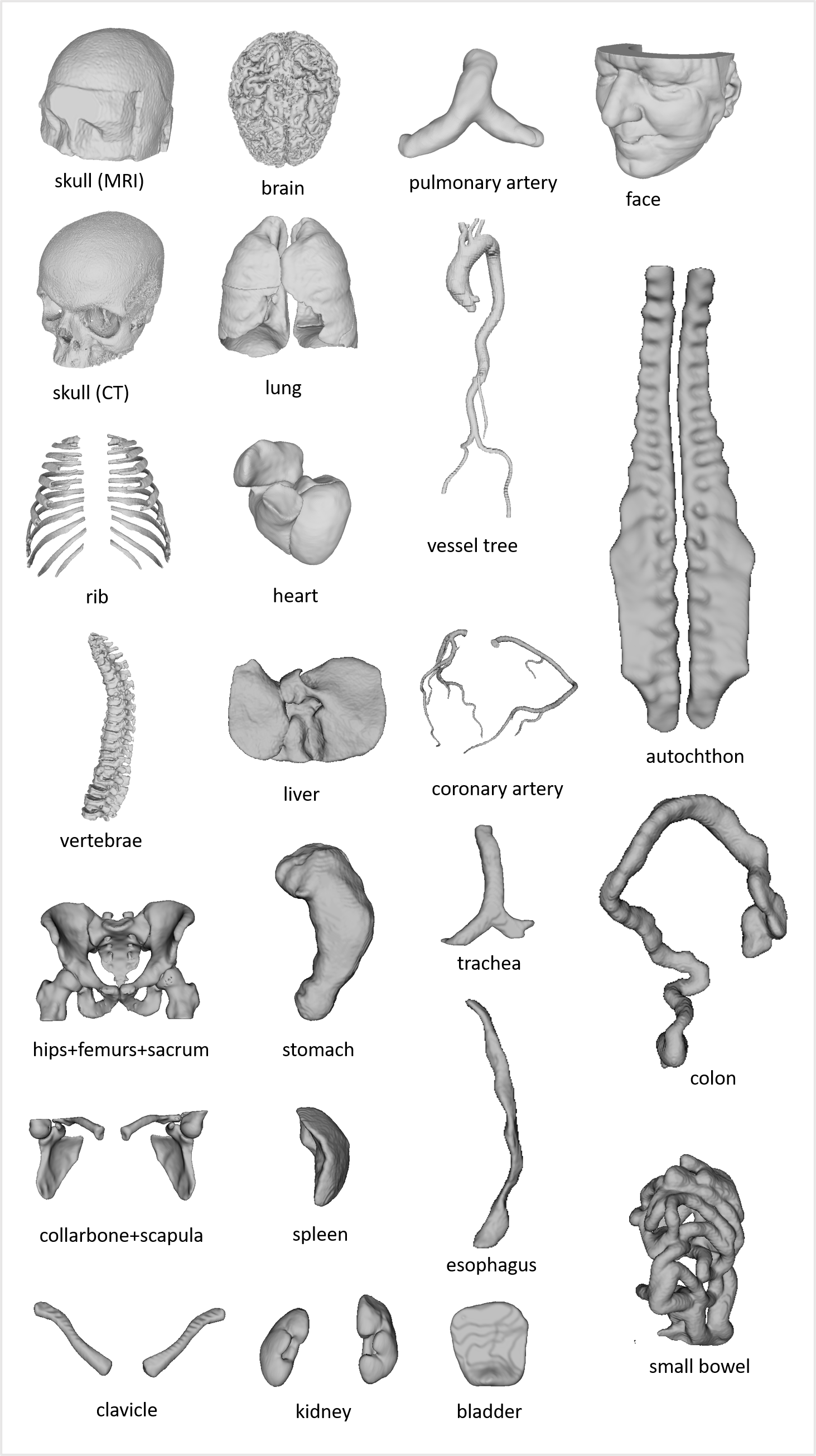}
 \caption{Example shapes in \textit{MedShapeNet}, including various bones (e.g., skulls, ribs and vertebrae), organs (e.g., brain, lung, heart, liver), vessels (e.g., aortic vessel tree and pulmonary artery) and muscles.}
\label{shape_gallery}
\end{figure}

While existing datasets, such as \textit{ShapeNet} are comprised of 3D computer-aided design (CAD) models of real-world objects (e.g., \textit{plane}, \textit{car}, \textit{chair}, \textit{desk}), \textit{MedShapeNet} provides 3D shapes extracted from the imaging data of real patients including healthy as well as pathological subjects (Fig. \ref{shape_gallery}).

\section{Shape and Voxel Features}
\label{shapevoxelfeatures}

Shapes describe objects’ geometries, provide a foundation for computer vision, and serve as a computationally efficient way to represent images despite not capturing voxel features. In medicine, numerous diseases alter the morphological attributes of the affected anatomical structures. For instance, neoplastic formations, such as tumors, significant alter the morphologies of organs like the brain and the liver (Fig. \ref{pathology}); Neurological disorders, including Alzheimer's disease (AD) \cite{ohnishi2001changes}, Parkinson's disease (PD) \cite{deng2022morphological} and substance use disorders, for instance, alcohol use disorder (AUD) and cocaine use disorder (CUD), can also cause morphological changes of brain substructures, such as the cerebral ventricles and the subcortical structures. These morphologic alterations allow disease detection and classification either manually, by medical professionals or automatically, through the application of specialised (e.g., shape analysis) machine learning algorithms. 

Hence, \textit{MedShapeNet} highlights the significance of shape features, including jaggedness, volume, elongation, etc., over voxel features, such as intensities, for disease characterization, current medical image analysis tasks are still dominated by voxel-based methods. For instance, the so-called voxel-wise spatial \textit{predictive maps}, as demonstrated by Akbari et al. \cite{akbari2016imaging}, can pinpoint areas of early recurrence and infiltration of glioblastoma. These maps can be effectively used for targeted radiotherapy \cite{seker2022tumor} (Fig. \ref{infiltrationmap}), as regions with high probability are associated with a greater risk of tumor recurrence after resection. A naturally arising question is whether such \textit{predictive maps} can be derived from the tumors' geometries. \textit{MedShapeNet} provides a platform to investigate the question and more:

\begin{figure}
\centering
\includegraphics[width=\linewidth]{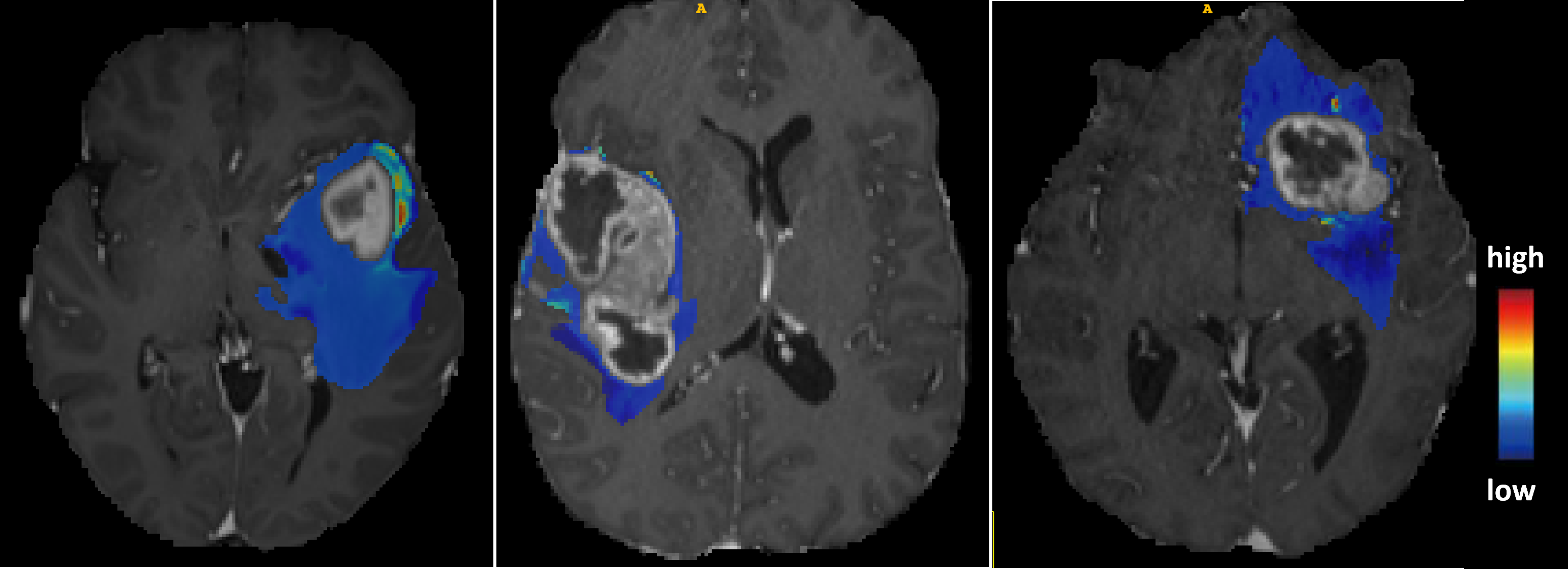}
 \caption{The \textit{predictive maps} overlaid onto patients' MRI scans. The \textit{predictive maps} are color-coded to indicate high or low probability of tumor infiltration.}
\label{infiltrationmap}
\end{figure}

\begin{itemize}
    \item What diseases can be comprehensively characterized by the shape features of the affected anatomical structures, and what diseases are solely reflected on voxel features? 
    \item How can one obtain discriminative shape features for disease detection using a machine learning model, either by handcrafting or learning them automatically using a deep network?
    \item How to effectively combine shape and voxel features when shape features alone are insufficient for disease detection? 
    \item Do changes in voxel and shape features correlate statistically, and if so, how can this correlation be quantified?
    \item Which of the current voxel-based mainstream approaches can be substituted with  computationally more efficient shape-based methods for the analysis of medical data?
\end{itemize}

Transitioning from gray-scale imaging data to shape data and shape-based methods brings three primary benefits:
\begin{enumerate}
\item Shape manifolds are spatially sparse, which enables the use of more computationally efficient algorithms, such as sparse convolutions \cite{li2023sparse}, point cloud \cite{jin2023ribseg} and mesh \cite{wickramasinghe2022weakly} neural networks; \item  Shape data contain less identifying information than gray-scale imaging data, reducing the vulnerability to privacy attack when they are publicly shared \cite{de2023guide}; 
\item Training on shape data encourages a deep network to concentrate on learning discriminative geometric features instead of patients' identities irrelevant to the task. This can help improve the robustness and trustworthiness and prevent identity-driven bias of the learning system.  
\end{enumerate}

\begin{table*}[ht]
\centering
\scriptsize
\caption{The Sources Segmentation Datasets (ordered alphabetically)}
\begin{tabular}[t]{llll} 
\toprule
Sources & Description & Dataset License\\ 
\hline
AbdomenAtlas~\cite{qu2023annotating} \href{https://github.com/MrGiovanni/AbdomenAtlas}{} & 25 organs and seven types of tumor & - \\
AbdomenCT-1K \cite{Ma2021AbdomenCT1K} \href{https://github.com/JunMa11/AbdomenCT-1K}{} & abdomen organs & \textbf{CC BY 4.0}\\
AMOS \cite{ji2022amos} \href{https://doi.org/10.5281/zenodo.7155725}{} &abdominal multi organs in CT and MRI & \textbf{CC BY 4.0}\\
ASOCA \cite{gharleghi2022automated,gharleghi2023annotated} \href{https://asoca.grand-challenge.org/}{}  &normal and diseased coronary arteries& -\\
autoPET \cite{jaus2023towards,gatidis2022whole,gatidis2023autopet,autopetdata} \href{https://autopet.grand-challenge.org/}{} & whole-body segmentations & \textbf{CC BY 4.0}\\
AVT \cite{radl2022avt} \href{https://doi.org/10.6084/m9.figshare.14806362}{} &aortic vessel trees &\textbf{CC BY 4.0} \\
BraTS \cite{baid2021rsna,menze2014multimodal,bakas2017advancing} \href{http://braintumorsegmentation.org/}{} & brain tumor segmentation & - \\
Calgary-campinas \cite{souza2018open} \href{https://portal.conp.ca/dataset?id=projects/calgary-campinas}{}  & brain structure segmentations &- \\
Crossmoda \cite{shapey2021segmentation,dorent2023crossmoda} \href{https://zenodo.org/record/6504722}{} &brain tumor and Cochlea segmentation  & \textbf{CC BY 4.0} \\
CT-ORG \cite{rister2020ct} \href{https://doi.org/10.6084/m9.figshare.13055663}{} & multiple organ segmentation & \textbf{CC0 1.0} \\
Digital Body Preservation \cite{vandenbossche2022digital} \href{https://opanex.discover.ilabt.imec.be/}{}  & 3D scans of anatomical specimens & - \\
EMIDEC \cite{lalande2020emidec,lalande2022deep}\href{https://emidec.com/segmentation-contest}{} & normal and pathological (infarction) myocardium & \textbf{CC BY NC SA 4.0}\\
Facial Models\cite{gsaxner2019facial} \href{https://doi.org/10.6084/m9.figshare.8857007.v2}{} & facial models for augmented reality  &\textbf{CC BY 4.0}\\
FLARE \cite{ma2022fast,simpson2019large,Ma2021AbdomenCT1K,FLARE22} \href{https://flare22.grand-challenge.org/}{} &13 Abdomen organs & -\\
GLISRT \cite{abcsdata,shusharina2021cross,shusharina2020automated} \href{https://doi.org/10.7937/TCIA.T905-ZQ20}{} &brain structures & \textbf{TCIA Restricted} \href{https://wiki.cancerimagingarchive.net/display/Public/Data+Usage+Policies+and+Restrictions}{}\\
\textit{HCP} \cite{elam2021human} \href{https://humanconnectome.org/}{} &paired brain-skull extracted from MRIs & \textbf{Data Use Terms}\href{https://www.humanconnectome.org/study/hcp-young-adult/document/wu-minn-hcp-consortium-open-access-data-use-terms}{} \\
HECKTOR \cite{andrearczyk2022overview,oreiller2022head} \href{https://hecktor.grand-challenge.org/}{} &head and neck tumor segmentation  &-\\
ISLES22 \cite{hernandez2022isles} \href{https://isles22.grand-challenge.org/}{} & ischemic stroke lesion segmentation & \textbf{CC-BY-4.0} \\
KiTS21 \cite{heller2020state} \href{https://github.com/neheller/kits21}{}  & kidney and kidney tumor segmentation & \textbf{MIT}\\
LiTS \cite{bilic2023liver} \href{https://competitions.codalab.org/competitions/17094}{} &liver tumor segmentation  &-\\
LNDb \cite{pedrosa2019lndb,pedrosa2021lndb} \href{https://lndb.grand-challenge.org/}{} &lung nodules & \textbf{CC BY NC ND 4.0}\\
LUMIERE \cite{suter2022lumiere} \href{https://doi.org/10.6084/m9.figshare.c.5904905.v1}{} & longitudinal glioblastoma & \textbf{CC BY NC} \\
\textit{MUG500+} \cite{li2021mug500} \href{https://doi.org/10.6084/m9.figshare.9616319}{} & healthy and craniotomy CT skulls & \textbf{CC BY 4.0}\\
MRI GBM \cite{Lindner2019} \href{https://doi.org/10.6084/m9.figshare.7435385.v2}{} & brain and GBM extracted from MRIs  & \textbf{CC BY 4.0} \\
PROMISE \cite{litjens2014evaluation} \href{https://zenodo.org/record/8014041}{} & prostate MRI segmentation  & - \\
PulmonaryTree~\cite{weng2023topology}  \href{https://github.com/M3DV/pulmonary-tree-repairing}{} & pulmonary airways, arteries and veins & \textbf{CC BY 4.0}\\
SkullBreak \cite{kodym2021skullbreak} \href{https://doi.org/10.6084/m9.figshare.14161307.v1}{} &complete and artificially defected skulls  & \textbf{CC BY 4.0} \\
SkullFix \cite{kodym2021skullbreak} \href{https://autoimplant2021.grand-challenge.org/Dataset/}{}  & complete and artificially defected skulls & \textbf{CC BY 4.0} \\
SUDMEX CONN \cite{angeles2022mexican}\href{https://openneuro.org/datasets/ds003346/versions/1.1.2}{} & healthy and (cocaine use disorder) CUD brains & \textbf{CC0}\\
TCGA-GBM \cite{bakas2017advancing} \href{https://www.nature.com/articles/sdata2017117}{}  & glioblastoma & -\\
3D-COSI \cite{gijs_luijten_2023_8379918} \href{https://zenodo.org/records/8379918}{} &3D medical instrument models & \textbf{CC BY 4.0} \\
3DTeethSeg \cite{ben20233dteethseg,ben2022teeth3ds} \href{https://github.com/abenhamadou/3DTeethSeg22_challenge}{} &3D Teeth Scan Segmentation  & \textbf{CC BY NC ND 4.0}\\
ToothFairy~\cite{cipriano2022improving,bolelli2023toothfairy} \href{https://toothfairychallenges.github.io/}{} &inferior alveolar canal& \textbf{CC BY SA}\\
\textit{TotalSegmentator} \cite{TotalSegmentator}  \href{https://doi.org/10.5281/zenodo.6802613}{} & various anatomical structures & \textbf{CC BY 4.0}\\
VerSe \cite{sekuboyina2020labeling} \href{https://github.com/anjany/verse}{} & large scale vertebrae segmentation & \textbf{CC BY 4.0} \\
\bottomrule
\end{tabular}
\label{table:quantitative_results}
\end{table*}

\section{Sources of Shapes}
\label{source_of_shapes}
\begin{figure}
\centering
\includegraphics[width=\linewidth]{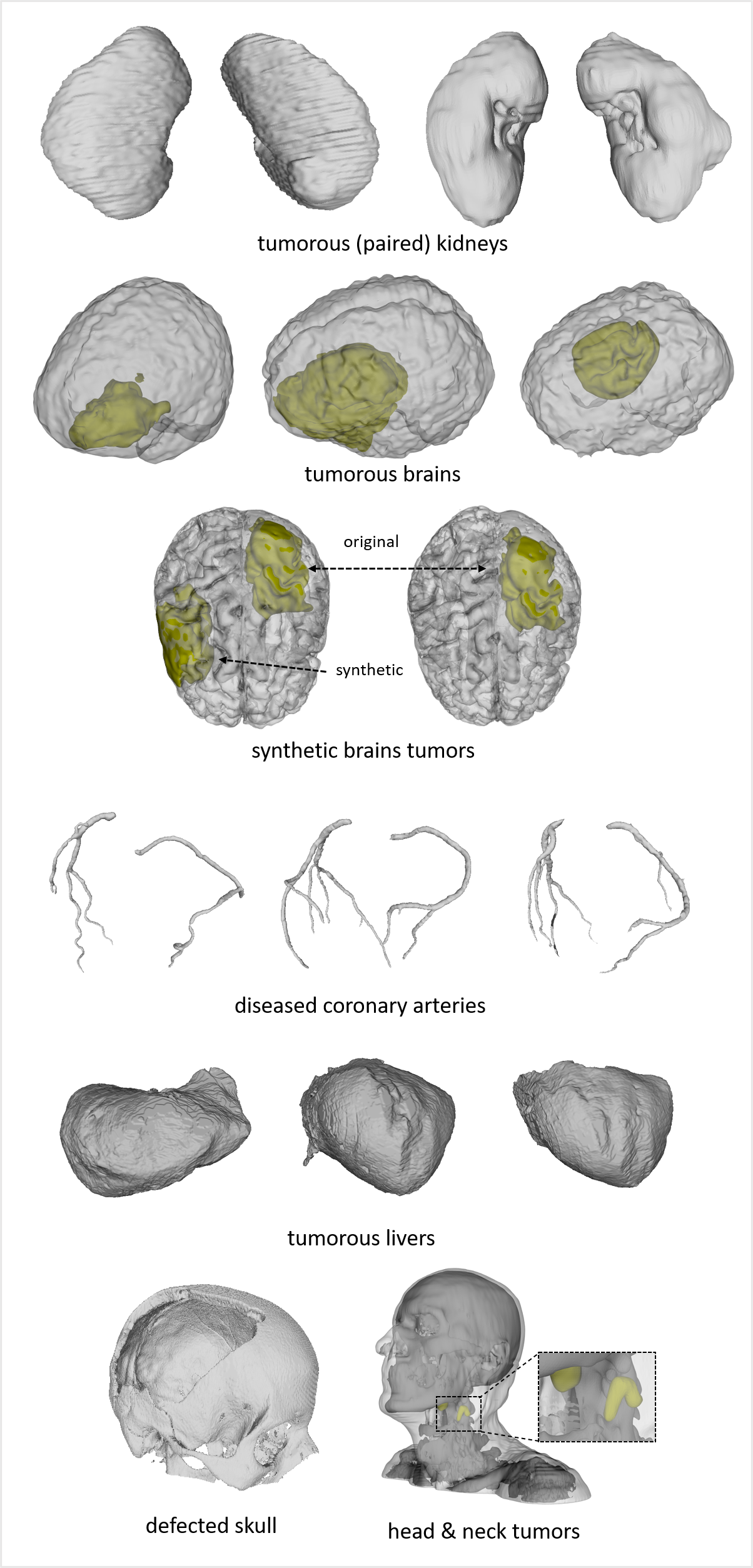}
 \caption{Example pathological shapes in \textit{MedShapeNet}, including tumorous kidney (paired), brain (with real and synthetic tumors), liver and head \& neck, as well as diseased coronary arteries. For illustration purpose, the opacity of some shapes is reduced to reveal the underlying tumors. We can study the effects of tumors on the morphological changes of an anatomy (e.g., brain) using such pathological data.}
\label{pathology}
\end{figure}

The shapes in \textit{MedShapeNet} mostly originate from high-quality segmentation masks of anatomical structures, including different organs, bones, vessels, muscles, etc. They are generated manually by domain experts, as those of the ground truth segmentations provided by medical image segmentation challenges \cite{EisenmannCVPR2023}, or semi-automatically, with the help of a segmentation network (e.g., \textit{TotalSegmentator}\cite{TotalSegmentator}, \textit{autoPET whole-body segmentation} \cite{jaus2023towards}, \textit{AbdomenAtlas}\cite{qu2023annotating}). The majority of semi-automatic segmentations were also quality-checked by experts. Anatomical shapes with sophisticated geometric structures, such as the pulmonary trees (Figure~\ref{pulmonary_trees}), are also included in the \textit{MedShapeNet} collection. In our terminology, we refer to binary voxel occupancy grids as segmentation masks, which we subsequently convert to meshes and point clouds using the \textit{Marching Cubes} algorithm \cite{lorensen1987marching}. The majority of the source segmentation datasets are Creative Commons (CC)-licensed (\autoref{table:quantitative_results}), allowing us to adapt and redistribute the data. Furthermore, \textit{MedShapeNet} includes both normal (Fig. \ref{shape_gallery}) and pathological shapes (Fig. \ref{pathology}), delivered by the imaging data of healthy and diseased subjects, respectively. In addition, \textit{MedShapeNet} provides 3D medical instrument models acquired using 3D handheld scanners \cite{gijs_luijten_2023_8379918} (Fig. \ref{medical_instrument}).

\subsection{AbdomenAtlas}\label{abdomenatlas}
The dataset provides masks of 25 anatomical structures and seven types of tumors, derived from 5,195 CTs of 26 hospitals across eight countries \cite{qu2023annotating}. These anatomical structures include the spleen, right kidney, left kidney, gall bladder, esophagus, liver, stomach, aorta, postcava, portal and splenic veins, pancreas, right and left adrenal glands, duodenum, hepatic vessel, right and left lungs, colon, intestine, rectum, bladder, prostate, left and right femur heads, and celiac trunk. Shape quality is ensured through manual annotations by medical professionals supported by a semi-automatic active learning procedure. The pathology-confirmed tumors include kidney, liver, pancreatic, hepatic vessel, lung, colon, and kidney cysts. The dataset provides a total of 51.8K tumor masks. Moreover, a novel modeling-based tumor synthesis method is used to generates small, synthetic ($<$20 mm) tumor shapes \cite{hu2023label,li2023early}]. 

\subsection{Pulmonary Trees} \label{pulmonarytree} 

\begin{figure*}
\centering
\includegraphics[width=\linewidth]{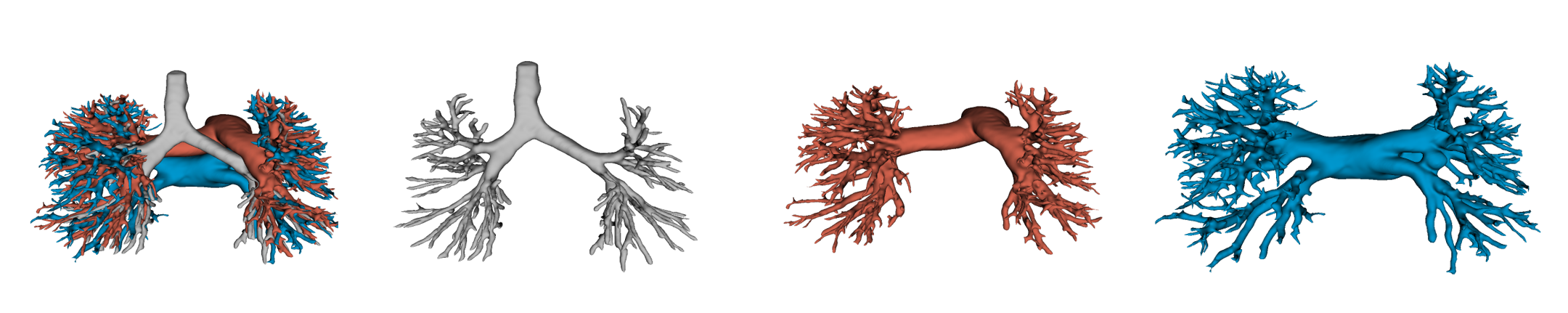}
 \caption{Illustration of a pulmonary tree comprising the airway, artery and vein---thin structures that are difficult to segment and reconstruct.}
\label{pulmonary_trees}
\end{figure*}

The PulmonaryTree dataset~\cite{weng2023topology} is a collection of pulmonary tree structures, amassed from 800 subjects across various medical centers in China~\cite{kuang2022makes}. It includes detailed 3D models of pulmonary airways, arteries, and veins, totaling $800\times 3=2,400$  shapes. Each 3D model originates from CT scans with  $512 \times 512$ voxels and 181 to 798 slices. The $Z$-spacing ranges from 0.5mm to 1.5mm. A collaborative annotation procedure ensures consistency provides a detailed and accurate representation of the pulmonary structures \cite{xie2023efficient}. This procedure required approximately \textbf{3 hours} per case.The PulmonaryTree dataset introduces complex tree-like structures, a challenging aspect in medical image analysis (Fig. \ref{pulmonary_trees}). Specific technical challenges include maintaining the continuity of thin structures and addressing the uneven thickness of the main and branch structures.

\subsection{TotalSegmentator} 
The dataset from Wasserthal et al. \cite{TotalSegmentator} includes over 1000 CT scans and the masks of 104 anatomical structures covering the whole body. The masks are generated automatically by a nnUNet \cite{isensee2021nnu}. The data have been used to improve diagnosis by correlating organ volumes with disease occurrences \cite{biomedinformatics3030036}.

\subsection{Human Connectome Projects (HCP)} 
The \textit{1200 Subjects Data Release} from HCP includes 1113 structural 3T head MRI scans of healthy young adults. From each scan, the \textit{Cortical Surface Extraction} script provided by \textit{BrainSuite} \footnote{\url{http://brainsuite.org/}} is used to extract the skull and brain masks. 

\subsection{MUG500+} 
This dataset contains the binary masks and meshes of 500 healthy human skulls and 29 craniectomy skulls with surgical defects \cite{li2021mug500}. Thresholding delivered the masks from head CT scans. 

\subsection{SkullBreak$/$SkullFix} 
The dataset includes the binary masks of healthy human skulls and the corresponding skulls with artificial defects.  Similar to \textit{MUG500+} \cite{li2021mug500}, thresholding head CTs from the \textit{CQ500} dataset \footnote{\url{http://headctstudy.qure.ai/dataset}} yields the masks.

\subsection{Aortic Vessel Tree (AVT)}
The dataset contains 56 computed tomography angiography (CTA) scans of healthy aortas and the masks of the aortic vessel trees \cite{radl2022avt}, including the aorta, the aortic arch, the aortic branch, and the iliac arteries (Fig. \ref{shape_gallery}).

\begin{figure*}
\centering
\includegraphics[width=0.9\linewidth]{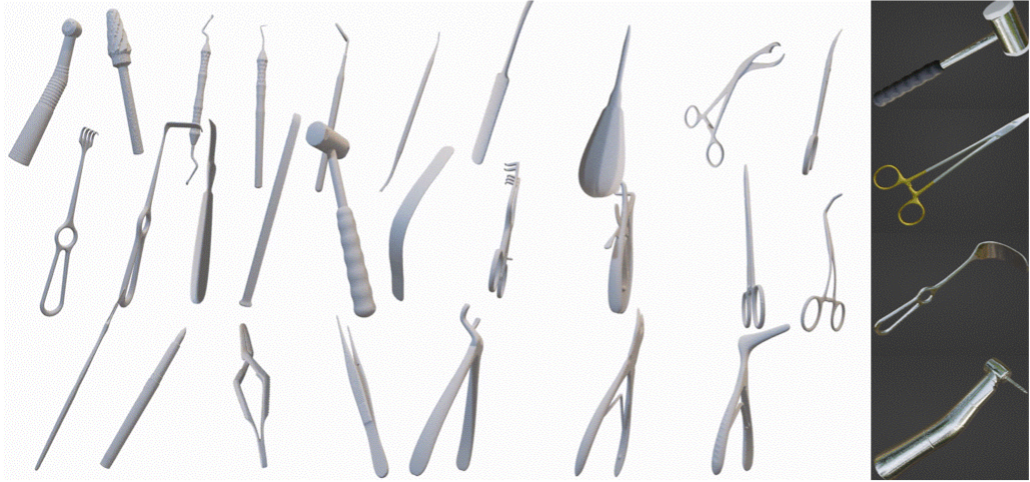}
 \caption{Illustration of 3D models of medical instruments used in oral and cranio-maxillofacial surgeries. The 3D models are obtained using structured light 3D scanners (Artec Leo from Artec3D and AutoScan Inspec from Shining 3D). Instrument models can be retrieved by the search query \textit{instrument} via the \textit{MedShapeNet} web interface. Image taken from \url{https://xrlab.ikim.nrw/}.}
\label{medical_instrument}
\end{figure*}

\subsection{Vertebrae Segmentation (VerSe)}
The \textit{VerSe} challenge provides the masks of vertebrae from around 210 subjects \cite{sekuboyina2020labeling}. In total, 2745 vertebra shapes are generated. 

\subsection{Automated Segmentation of Coronary Arteries (ASOCA)}
The \textit{ASOCA} challenge provides the manual segmentations of 20 normal and 20 diseased coronary arteries \cite{gharleghi2023annotated}.

\subsection{3D Teeth Scan Segmentation and Labeling Challenge (3DTeethSeg)}
Automated teeth localization, segmentation, and labeling from intra-oral 3D scans significantly improve dental diagnostics, treatment planning, and population-based studies on oral health. Before initiating any orthodontic or restorative treatment, it is essential for a CAD system to accurately segment and label each instance of teeth. This eliminates the need of time-consuming manual adjustments by the dentist. The \textit{3DTeethSeg} provides the upper and lower jaw scans of 900 subjects, and the manual segmentations of the teeth, obtained from clinical evaluators with more than 10 years of expertise \cite{ben20233dteethseg,ben2022teeth3ds}.

\subsection{Lung Cancer Patient Management (LNDb) Challenge}
This dataset comprises lung nodule in low-dose CTs recorded for lung cancer screening \cite{pedrosa2019lndb,pedrosa2021lndb}. A total of 861 lung nodule masks correspond to 625 individual nodules segmented from 204 CTs. Five radiologists identified all pulmonary nodules with an in-plane dimension of 3 millimeters and higher. 

\subsection{Evaluation of Myocardial Infarction from Delayed-Enhancement Cardiac MRI (EMIDEC)}
This \textit{EMIDEC} challenge provides 150 delayed enhancement MRI (DE-MRI) images in short axis orientation of the left ventricles. Experts contoured the myocardium and infarction areas in normal (50 cases) and pathological (100 cases) cases \cite{lalande2020emidec,lalande2022deep}. The images were acquired roughly 10 minutes after the injection of a gadolinium-based contrast agent. The dataset is owned by the University Hospital of Dijon (France), but it is freely available. 

\subsection{ToothFairy}
Placing dental implant can become complex when the implant hits the inferior alveolar nerve. The \textit{ToothFairy} dataset contains cone-beam computed tomography (CBCT) images and was released for a segmentation challenge in 2023 \cite{bolelli2023toothfairy}. It extends the previous datasets (i.e., \cite{dibartolomeo2023inferior}) and comprises 443 dental scans with a voxel size of $0.3$\,mm$^3$ yielding volumes with shapes ranging from $(148, 265, 312)$ to $(169, 342, 370)$ across the Z, Y, and X axes, respectively. The dataset includes 2D sparse annotations for all 443 volumes, while only a subset of 153 volumes contains detailed 3D voxel-level annotations. A team of five experienced surgeons delivered the ground truth \cite{lumetti2023annotating,mercadante2021cone}. Additionally, a test set of 50 CBCT with a voxel size of $0.4$\,mm$^3$ is provided for evaluation.

\subsection{HEad and neCK TumOR segmentation and outcome prediction (HECKTOR)}
The training set of the \textit{HECKTOR} challenge comprises 524 PET-CT volumes from seven hospitals with manual primary tumor and metastatic lymph nodes contours \cite{andrearczyk2022overview}. The data originates from FDG-PET and low-dose non-contrast-enhanced CT images of the head and neck region of subjects suffering from oropharyngeal cancer.   The training set of the this challenge is provided to \textit{MedShapeNet}.

\subsection{autoPET}
Similar to \textit{TotalSegmentor}, whole-body segmentations are extracted from the PET-CT dataset provided by the \textit{autoPEt} challenge \cite{gatidis2022whole}, using an semi-supervised segmentation network \cite{jaus2023towards}. The dataset comes from cancer patients and includes manual masks of tumor lesions. 

\subsection{Calgary-Campinas (CC)}
This dataset provides high-quality anatomical data with $1$\,mm$^3$ voxels from T1-weighted MRIs of 359 healthy subjects on scanners from three different vendors (GE, Philips, Siemens) at field strengths of 1.5 T and 3 T \cite{souza2018open}. The subjects vary in age and gender (176 M: 183 F, 53.5 +/- 7.8 years, min:18 years, max: 80 years). Probabilistic brain masks resulted from eight automated brain segmentation algorithms by simultaneous truth and performance level estimation (STAPLE) \cite{warfield04}. The quality of the masks was validated against 12 manual brain segmentations. Scientists investigate brain extraction models \cite{lucena2019convolutional}, domain shift and adaptation in brain MRI \cite{saat2022domain}, as well as MRI reconstruction \cite{yiasemis2022recurrent} using the CC dataset. 

\subsection{Abdominal Multi-Organ Benchmark for Segmentation (AMOS)}
The AMOS data includes 500 CTs and 100 MRIs from a variety of scanners and locations \cite{ji2022amos}. It provides expert segmentations of 15 abdominal organs: spleen, right kidney, left kidney, gallbladder, esophagus, liver, stomach, aorta, inferior vena cava, pancreas, right adrenal gland, left adrenal gland, duodenum, bladder, and prostate/uterus. Patients with abdominal tumors or other abnormalities delivered the images. 

\subsection{AbdomenCT-1K and Fast and Low-resource Abdominal Organ Segmentation (FLARE)}
This dataset includes more than 1000 CTs and manually generated masks of the liver, kidney, spleen, and pancreas \cite{Ma2021AbdomenCT1K}. A subset of the dataset was used in the \cite{FLARE} challenge, which provides expert segmentations of 13 abdomen organs the right and left kidney, stomach, gallbladder, esophagus, aorta, inferior vena cava, right adrenal gland, left adrenal gland, and duodenum \cite{ma2022fast}. some of the CT scans are acquired from cancer patients. 

\subsection{Ischemic Stroke Lesion Segmentation (ISLES)}
The \textit{ISLES} challenge \cite{hernandez2022isles} provides 250 brain MRIs with binary masks depicting stroke infarctions. The dataset encompasses diverse brain lesions in terms of volume, location, and stroke pattern. Masks are generated by manually refining automatic segmentations from a 3D UNet \cite{cciccek20163d}.

\subsection{Synthetic Anatomical Shapes and Shape Augmentation}
In addition to real anatomical shapes, we also provide synthetic shapes generated by generative adversarial networks (GANs) \cite{ferreira2022gan}. For instance, we generate synthetic tumors for 27,390 real brains (Fig. \ref{pathology}). Besides GANs, synthetic shapes can also be generated by registering two shapes and warping them to each other’s spaces \cite{ellis2020deep}. This registration-based shape augmentation methods were used in the winning solutions of both the \textit{AutoImplant I} and \textit{AutoImplant II} challenges \cite{li2023towards,li2021autoimplant}. 

\subsection{Medical Instruments}
In addition to anatomical shapes,  \textit{MedShapeNet} also provides 3D models of medical instruments \cite{gijs_luijten_2023_8379918}, such as drill bits, scalpels, and chisels (Fig. \ref{medical_instrument}). We process the structured-light 3D scans using proprietary software (Ultrascan 2.0.0.7, Artec Studio 17 Professional) to remove noise. These models could help develop surgical tool tracking methods in mixed reality for medical education and research \cite{gsaxner2021inside}. Realistic and accurate virtual surgical planning is performed in AR or VR \cite{velarde2023virtual}, which improves the surgical outcome \cite{laskay2023optimizing}.

\subsection{Digital Body Preservation Repository}
These 3D models were captured from anatomical specimens using the handheld, high-resolution (accuracy 0.05 mm) structured-light surface scanner (Space Spider) and processed by the Studio 15 software (Artec 3D LUX, Luxembourg, Luxembourg) \cite{vandenbossche2022digital}. 

\begin{figure*}
\centering
\includegraphics[width=0.9\linewidth]{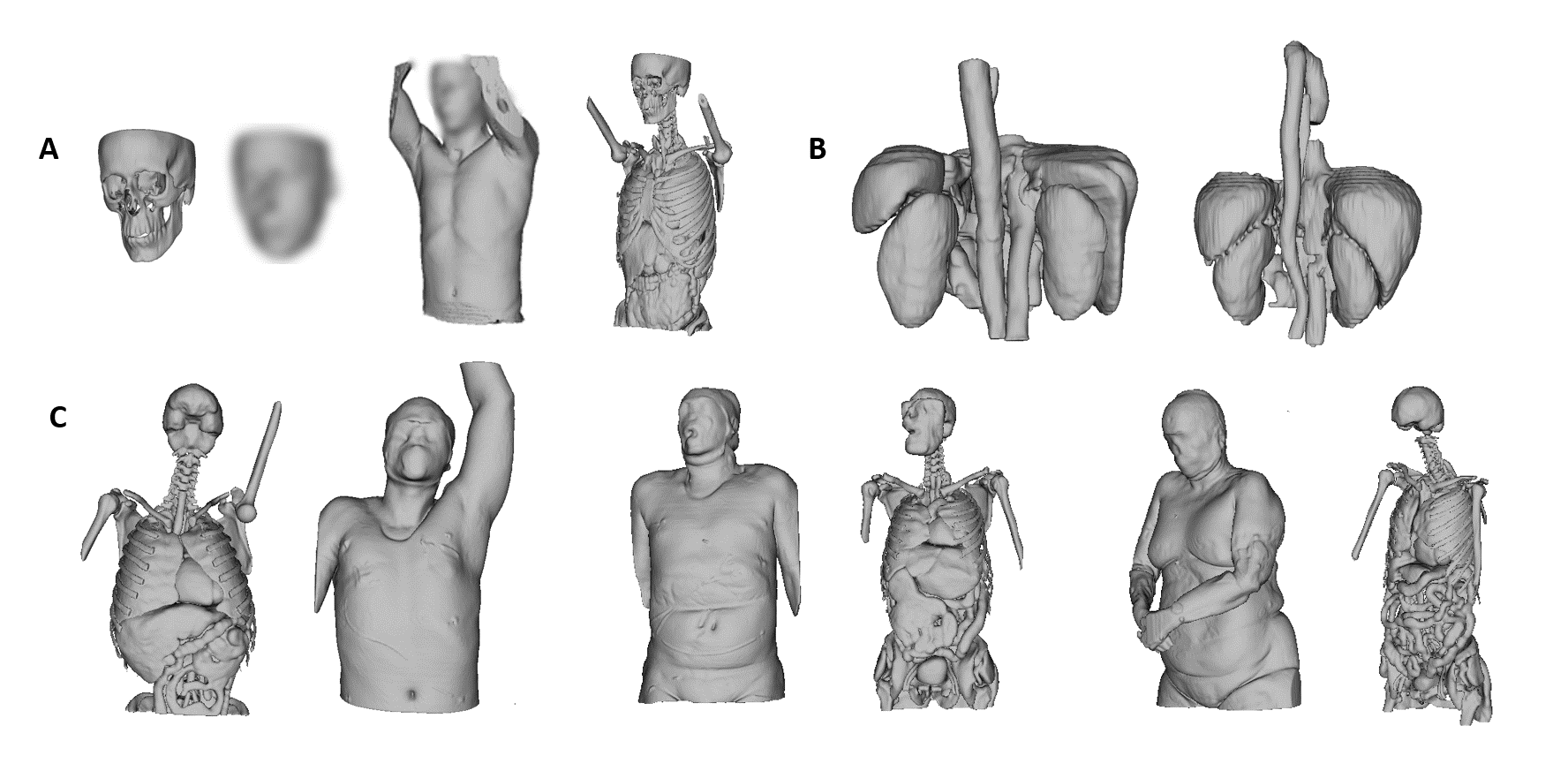}
 \caption{Examples of paired anatomical shapes in \textit{MedShapeNet}. (A) Paired skins, muscles, fat, different tissues, organs and bones. (B) Paired abdominal anatomies, including liver, spleen, pancreas, right kidney, left kidney, stomach, gallbladder, esophagus, aorta, inferior vena cava, right adrenal gland, left adrenal gland, and duodenum. (C) Paired internal anatomies and body surfaces. For anonymity, the faces are blurred.}
\label{annotation}
\end{figure*}

\subsection{Pathological Shapes}
\label{pathologicalshapes}
To increase the variability of the shape collections, \textit{MedShapeNet} contains not only normal/healthy anatomical shapes, such as the kidneys from \textit{TotalSegmentor} and the brains from \textit{HCP}, but also pathological ones, which are derived from patients diagnosed with a specific pathological condition, such as tumor (liver, kidney, etc) and CUD (SUDMEX CONN, Table \ref{table:quantitative_results}). Fig. \ref{pathology} shows the tumorous kidneys, brains, livers and head \& neck, as well as diseased coronary arteries from different sources. We also use generative adversarial networks (GANs) to generate synthetic brain tumors, as shown in Fig. \ref{pathology}.

\section{Annotation and Example Use Cases}
\label{annotation_section}
In \textit{MedShapeNet}, \textit{pairedness} is defined as having two composites i.e., the anatomical shapes and the metadata originating from the same subject, with one serving as input and the other as the ground truth. For instance, a 3D shape in \textit{MedShapeNet} is paired with its anatomical category, such as ’liver’, ’heart’, ’kidney’, and ’lung’, which can be used for anatomical shape classification and retrieval. The metadata from DICOM or medical reports provides precise information about the source images, the patients (including attributes such as gender, age, body weight) as well as the diagnosis, and can deliver a variety of annotations. Synthetic shapes are distinguished from those obtained from real imaging data by the ‘synthetic’ label.

\subsection{Benchmarks Derived from \textit{MedShapeNet}}

From \textit{MedShapeNet} and its paired data, we can derive three types of benchmark datasets (Table \ref{table:paired}):

\begin{table*}[ht]
\centering
\scriptsize
\caption{Instances of \textit{MedShapeNet} Benchmarks (abbr. AD - Alzheimer's disease, AUD - alcohol use disorder, CUD - cocaine use disorder)}
\begin{tabular}[t]{ll|ll|ll} 
\toprule
\multicolumn{2}{c}{\textit{Discriminative Benchmarks}} & \multicolumn{2}{c}{\textit{Reconstructive Benchmarks}} & \multicolumn{2}{c}{\textit{Variational Benchmarks}} \\
\hline
input (shape) & ground truth (metadata) & input (shape) & ground truth (shape) & input (shape+metadata) & ground truth (shape)\\

\hline
liver/kidney/brain & tumor/healthy &skull &face & face + AUD/CUD/AD/age & face \\
brain & AUD/CUD/AD/age  & ribs+spines & torso organs & brain + AUD/CUD/AD/age & brain\\
face & AUD/CUD/age/gender & skin& body fat/muscle/skeleton & - & - \\
3D shapes& anatomical categories &full skeleton &skin & - & - \\
\bottomrule
\end{tabular}
\label{table:paired}
\end{table*}

\begin{itemize}
\item \textbf{Discriminative benchmarks} are comprised of 3D shapes and the corresponding anatomical categories and diagnosis. They can be used to train a classifier to discriminate 3D shapes (e.g., healthy, cancerous) based on shape-related features.
\item \textbf{Reconstructive benchmarks} are composed of anatomical shapes derived from whole-body segmentations. They can be used in shape reconstruction tasks. For example, by training on paired skull-face shapes (Fig. \ref{annotation} (A)), we can reconstruct human faces from the skulls automatically. We can also estimate an individual's body composition, such as fat percentage or muscle distribution from the body surface \cite{mueller2023body,piecuch2023muscle}, by regressing on paired skin-fat or skin-muscle data (Fig. \ref{annotation} (C)), or create a missing organ from its surrounding anatomies \cite{li2023completor}.
\item \textbf{Variational benchmarks} are usually used for conditional reconstruction of 3D anatomical shapes. In addition to the geometric constraints imposed by the input shape, new reconstructions are expected to satisfy an additional attribute, such as age, gender or pathology. For example, it is possible to reconstruct multiple faces of different ages from the same skull, by introducing \textit{age} as a constraint during supervised training. Similarly, a pathological condition, such as tumor, can be imposed on healthy anatomies, or the morphological changes of an anatomy during disease progression can be modeled \cite{sauty2022progression}. Variational auto-encoder (VAE) \cite{kingma2013auto} and GANs are commonly used for such conditional reconstruction tasks.
\end{itemize}

\begin{figure*}
\centering
\includegraphics[width=0.9\linewidth]{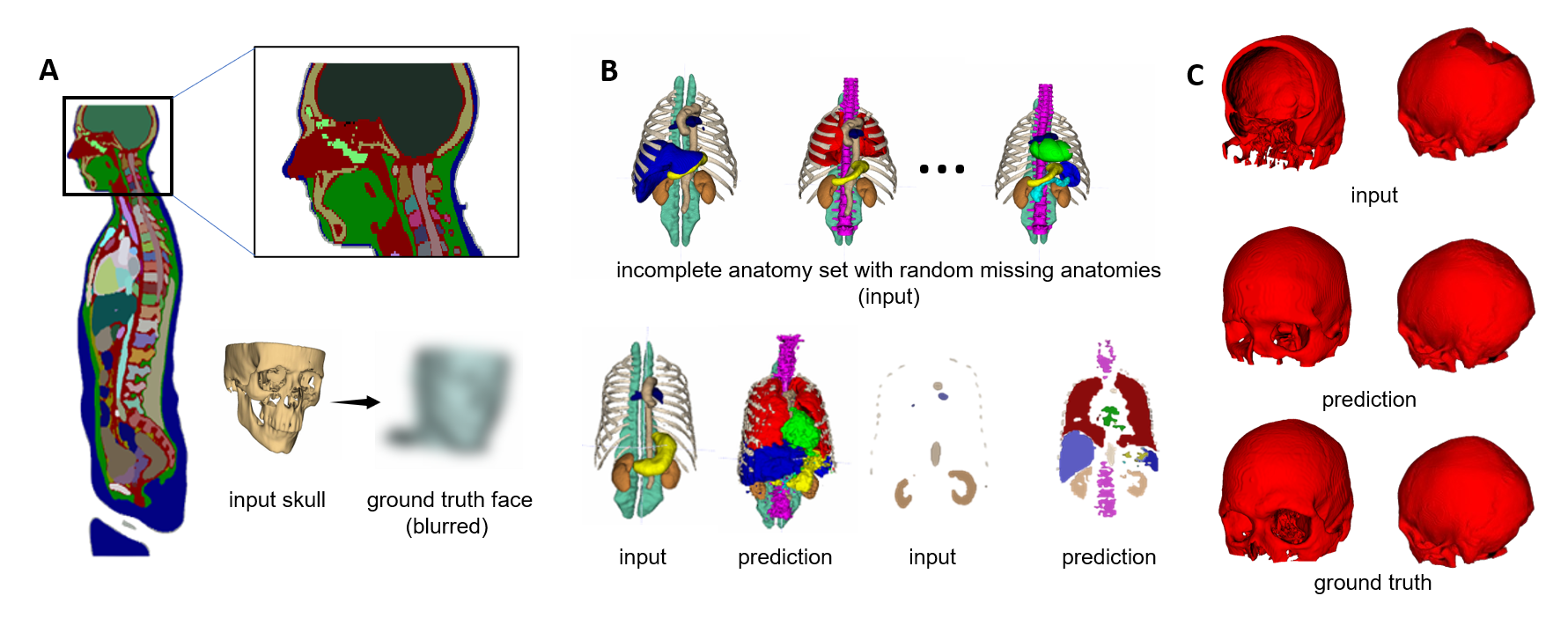}
 \caption{Benchmarks for various vision applications can be derived from \textit{MedShapeNet}, such as (A) forensic facial reconstruction, (B) anatomical shape reconstruction, and (C) skull reconstruction.}
\label{use_cases}
\end{figure*}

\subsection{Example Use Cases of \textit{MedShapeNet}}
\label{usecase}
\begin{figure*}
\centering
\includegraphics[width=0.9\linewidth]{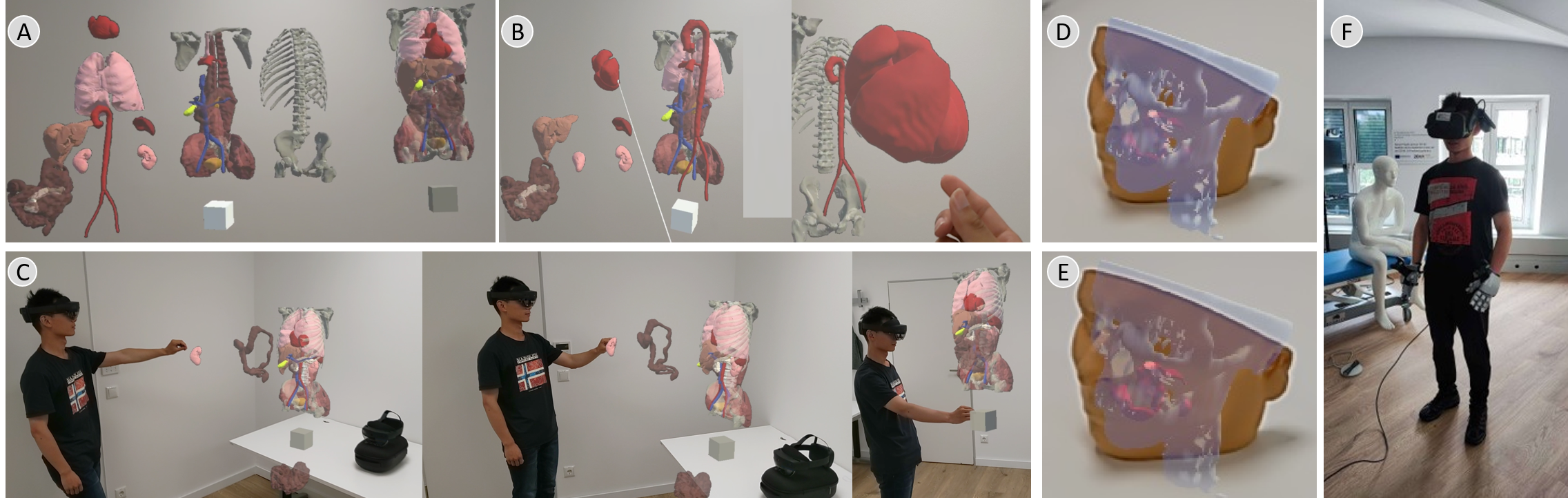}
 \caption{A use case of \textit{MedShapeNet} in AR- and VR-based anatomy education. (A) a whole-body model from \textit{MedShapeNet} dissembled into individual anatomies. (B, C) anatomy manipulation in first- and third-person views. (D, E) a 3D-printed facial phantom and the corresponding skull and tumors. (F) using haptic VR gloves to interact with the 3D anatomical models in the virtual environment.}
\label{arusecase}
\end{figure*}

To illustrate the unique value of \textit{MedShapeNet}, we describe five real-world use cases and show how \textit{MedShapeNet} is used to solve vision/medical problems:
\begin{itemize}
\item \textbf{Tumor classification} of brain lesions is usually based on gray-scale MRIs  \cite{amin2019brain,amin2021brain}. In this use case, we train a convolutional neural network (CNN)-based classifier to discriminate between tumorous and healthy brain shapes. The classifier has shown good convergence and generalizability. Similar results are observed for the classification of brain shapes from males and females, in line with existing studies \cite{xin2019brain}. 

\item \textbf{Facial reconstruction} is a common practice in archaeology, anthropology and forensic science, where the objective is to recreate the facial appearances of historical figures, ancient humans or victims from their skeletal remains \cite{missal2023forensic}. Orthognathic surgery also employs this technology to predict postoperative outcomes \cite{lampen2023spatiotemporal}. Nevertheless, in addition to the skull, the facial appearance is also significantly influenced by factors such as the quantity and distribution of facial fat and muscles \cite{damas2020relationships}, making facial reconstruction a highly ill-posed problem in terms of the skull-face relationship (Fig. \ref{use_cases} A).

\item \textbf{Skull Reconstruction} aims to rebuild missing parts of the skull bones around the facial area or the cranium (Fig. \ref{use_cases} (C)), where both voxel grids \cite{li2022training,li2023towards,li2021autoimplant} and point clouds \cite{friedrich2023point,wodzinski2023high} have been used to represent the skull data.

\item \textbf{Anatomy completion} investigates the feasibility of automatically generating whole-body segmentations given only sparse manual annotations. The generated segmentations can subsequently be used as pseudo labels to train a whole-body segmentation network \cite{li2023completor}. Fig. \ref{use_cases} (B) provides an example input and the corresponding reconstruction results.

\item \textbf{Extended reality (XR)} combines real and virtual worlds. \textit{MedShapeNet} can also benefit a variety of XR (AR/MR/VR) applications that require 3D anatomical models \cite{gsaxner2023hololens}, such as  virtual anatomy education \cite{bolek2021effectiveness}. Fig. \ref{arusecase} (A) shows a whole-body model using the \textit{Microsoft HoloLens} AR glasses. The user can dissemble individual anatomies, move them, zoom in and out, and rotate the structures (Fig. \ref{arusecase} (B) and Fig. \ref{arusecase} (C)). Furthermore, if necessary, we can 3D print the models (Fig. \ref{arusecase} (D) and Fig. \ref{arusecase} (E)). Users can also wear VR gloves (Fig. \ref{arusecase} (F)) to receive haptic feedback while interacting with the 3D anatomies in VR \cite{hapticKathrin}. 
\end{itemize}

\section{\textit{MedShapeNet} Interface}
\label{webinterface}

Two interfaces are created for \textit{MedShapeNet}, including a web-based interface that provides access to the original high-resolution shape data, and a Python API that enables users to interact with the shape data via Python.

\subsection{Web-based Interface}

A user-centric, intuitive web-based interface \footnote{\url{https://medshapenet.ikim.nrw/}} has been developed to provide convenient access to the shape data within \textit{MedShapeNet}, which allows users to search, retrieve, and view individual shapes. Shapes can be retrieved using queries related to anatomical category such as \textit{heart'}, \textit{brain}, \textit{hip}, \textit{liver}, or pathologies like \textit{tumor}. A dedicated GitHub page \footnote{\url{https://github.com/Jianningli/medshapenet-feedback}} has also been established to manage shape contribution and removal (in case of inaccurate shapes), feature requests and the open-sourcing of applications based on \textit{MedShapeNet}.

\subsection{\textit{MedShapeNetCore} and Python API}
\label{MedShapeNetCore}
We have also developed a Python API that facilitates the integration of the dataset into Python-centric workflows for computer vision and machine learning. This API grants access to a standardized subset of the original \textit{MedShapeNet} dataset, referred to as \textit{MedShapeNetCore}, which has been specifically curated for the efficient and reliable benchmarking of various vision algorithms. \textit{MedShapeNetCore} differs from the original dataset in aspects:

\begin{itemize}
    \item \textbf{Resolution.} The original 3D models are prohibitively high resolution to be used directly by vision algorithms \footnote{The typical resolution for segmentation masks is  $512\times512\times Z$, which corresponds to hundreds of thousands points in point representations. Dense anatomical structures such as the brain typically contain several million points.}. In contrast, \textit{MedShapeNetCore} contains considerably more lightweight 3D models and lower resolution images, similar to those in \textit{ShapeNet} \cite{chang2015shapenet} and \textit{MedMNIST} \cite{yang2023medmnist}.
    \item \textbf{Quality.} The 3D models in \textit{MedShapeNetCore} are water-tight and the quality of each individual model has been meticulously verified through manual inspection.
    \item \textbf{Annotation.} \textit{MedShapeNetCore} is more densely annotated, expanding its applicability to tasks such as shape part segmentation \cite{wang2015semantic} and anatomical symmetry plane estimation.
\end{itemize}

The 3D shapes are stored in the standard formats for geometric data structures, i.e., NIfTI (.nii) for voxel grids, stereolithography (.stl) for meshes and Polygon File Format (.ply) for point clouds, facilitating fast shape preview via existing softwares. The Python API facilitates the loading of these shape data into standard \textit{Numpy} arrays, ensuring a seamless transformation into tensor representations compatible with various deep learning frameworks, including but not limited to \textit{PyTorch}, \textit{MONAI}, and \textit{TensorFlow}. The light-weight nature of these data expedites the process of developing new medical vision algorithms or evaluating existing ones, while maintaining a low computational overhead. The ongoing efforts in the development of the Python API include integrating PyTorch3D \cite{ravi2020accelerating} to leverage its sophisticated 3D operators, establishing predefined benchmarks tailored for various vision and medical applications, and incorporating pre-trained models and shape processing algorithms.

\section{Discussion}
\label{discussion_futurework}
High-quality, annotated datasets are valuable assets for data-driven research. We created MedShapeNet as an open, ongoing effort and requires continuous contributions from these communities.   
We believe that \textit{MedShapeNet} holds the potential to make significant contributions to research in medical imaging and computer vision. It could impact the practice of medical data curation and sharing, as well as the development of data-driven methods for medical applications. 

Compared to vision datasets, large medical datasets are more difficult to curate due to the sensitive, distributed, and scarce nature of medical images. Therefore, the medical imaging community has recently started catching up with the development of vision algorithms that can exploit large datasets, with more and more medical researchers becoming open to data-sharing. Thus, \textit{MedShapeNet} provides a versatile dataset that both vision and medical researchers are accustomed to.

To avoid potentially harmful societal impact, computer vision research involving human-derived data should be conducted with care. We designed MedShapeNet specifically for research, and the researchers shall follow ethical guidelines throughout methodology development and experimental design. For example, publicly sharing neuroimaging data bears high privacy risks and needs regulation, since they contain patients’ facial profiles \cite{khalid2023privacy}. For instance, Schwarz et al. recently identified participants in a clinical trial comparing their faces reconstructed from MRI with photographs on social media \cite{schwarz2019identification}. Therefore, besides removing patients’ meta informationfrom DICOM tags, defacing is also commonly practiced \cite{giessler2021facial}. However, we have shown that machine learning can reconstruct skulls even when they are damaged or parts of the bones are missing. Another double-edged use case of \textit{MedShapeNet} is training machine learning to detect substance (drug or alcohol) addiction or other diseases e.g., fetal alcohol syndrome (FAS), based on facial characteristics \cite{mclaughlin2010interactive}. Furthermore, since \textit{MedShapeNet} preserves the correspondence between the shapes and patients’ meta information, such as age, race, gender, medical history, etc., which facilitates the learning of some controversial mapping relationships. Potentially, the ethnic identity or medical history is predicted from a person’s skull or facial profiles \cite{suzuki2020examination}. It is therefore the responsibility of the researchers to weigh the social benefits against the potential negative societal impacts while developing models using \textit{MedShapeNet}.

For future developments, we will primarily focus on the following aspects:

\begin{itemize}
    \item Incorporating a greater number of datasets and metadata as well as pathological shapes, particularly those pertaining to rare diseases.
    \item Advocating for \textit{MedShapeNet} through presentations at conferences, symposia, and seminars, as well as organizing hackweeks, workshops, and challenges. 
    \item Establishing additional benchmarks and use cases.
    \item Enhancing the web and Python interfaces.
\end{itemize}

\section{Conclusion}
In this white paper, we have introduced the initial efforts for MedShapeNet. We (1) formed a community for data contribution; (2) derived open-source benchmark datasets for several use cases; (3) constructed interfaces to search to download the shape data and its paired information; (4) brought up several interesting shape-related research topics; and (5) discussed the relevance of ethical guidelines and precautions for privacy of medical data.

\section*{Acknowledgments}
This work was supported by the REACT-EU project KITE (Plattform für KI-Translation Essen, EFRE-0801977, \url{https://kite.ikim.nrw/}), FWF enFaced 2.0 (KLI 1044, \url{https://enfaced2.ikim.nrw/}), AutoImplant (\url{https://autoimplant.ikim.nrw/}) and „NUM 2.0“ (FKZ: 01KX2121). Behrus Puladi was funded by the Medical Faculty of RWTH Aachen University as part of the Clinician Scientist Program. In addition, we acknowledge the National Natural Science Foundation of China (81971709; M-0019; 82011530141). The work of J. Chen was supported by the Bundesministerium für Bildung und Forschung (BMBF, Ref. 161L0272). The work of ISAS was supported by the “Ministerium für Kultur und Wissenschaft des Landes Nordrhein-Westfalen” and “Der Regierende Bürgermeister von Berlin, Senatskanzlei Wissenschaft und Forschung”. Furthermore, we acknowledge the \textit{Center for Virtual and Extended Reality in Medicine} (ZvRM, \url{https://zvrm.ume.de/}) of the University Hospital Essen. The CT-ORG dataset was obtained from the Cancer Imaging Archive (TCIA). CT-ORG was supported in part by grants from the National Cancer Institute, 1U01CA190214 and 1U01CA187947. We thank all those who have contributed to the \textit{MedShapeNet} collection (directly or indirectly).

\ifCLASSOPTIONcaptionsoff
  \newpage
\fi

\bibliographystyle{IEEEtran}
\bibliography{references}

\end{document}